\documentclass[11pt,a4paper]{article}

\usepackage[utf8]{inputenc}
\usepackage[T1]{fontenc}
\usepackage[english]{babel}
\usepackage{amsmath,amssymb}
\usepackage{graphicx}
\usepackage{booktabs}
\usepackage{hyperref}
\usepackage{geometry}
\usepackage{microtype}
\usepackage{placeins}

\DeclareRobustCommand{\placeholder}{\ifmmode\text{\ttfamily\textless\_\_\_\_\_\_\textgreater}\else{\ttfamily\textless\_\_\_\_\_\_\_\textgreater}\fi}

\title{LoRA Direction Extraction for Controllable Light Toggling in FLUX.1 Kontext}

\author{
{\Large\bfseries BURO.io Research Lab}
\\[0.25em]
{\small\url{https://buro.io}}
\\[0.8em]
Petr Golenderov, Dmitry Mazyar, Natalia Sovpel, Alexander Aksenov
\\[0.8em]
\hfill
\parbox{0.5\textwidth}{
  \raggedleft
  \small
  We express our gratitude to Maxim Korytko and Sviatoslav Pankratov for their assistance in organizing the inference process.
}
}

\date{}

\begin{document}

\maketitle

\begin{abstract}
We propose a fine-tuning method for flow-matching diffusion models aimed at realistic artificial light modeling without the need for a large training dataset. We address the task of controllable interior image editing, where the goal is to turn artificial light sources on or off while preserving the scene geometry, object placement, materials, and visual identity of the original image. To achieve this, we decompose the task into two independent formulations. We introduce the LoRA Direction Training Method, which extracts the pure direction of the LoRA adapter effect in the diffusion model flow field, and we also introduce specialized loss functions to ensure the realism of the inverse transformation. Additionally, the resulting increment map is used for more precise adjustment of the lighting color and temperature.
\end{abstract}

\section{Introduction}
Modern image editing models demonstrate high quality across a wide range of transformations, including object insertion, style transfer, and global scene modification. However, artificial lighting control remains a challenging task. Altering the lighting can significantly transform an interior image. A model must not merely change the appearance of a lamp but also update the associated lighting effects in a physically and visually plausible manner. Turning on a lamp affects not only the local brightness near the light source but also the distribution of shadows, reflections, color tones, and the overall visual balance of the scene. Similarly, turning off the artificial light should result in a consistent change in both global and local illumination. At the same time, any unintended changes to the composition, object shapes, or materials are perceived as a disruption of the interior identity. Consequently, editing the lighting states is a more complex task than standard local editing based on text prompts.

In this work, we approach the task of artificial light activation by treating the operation not as the addition of any new content, but as a change of the state of the already-existing scene — a concept formally akin to Style Transfer~\cite{gatys2016styletransfer}. Activating a light can be viewed as a directed transformation of the visual state of the scene: the light source is already present in the image, but its activity level and the resulting illumination distribution change. This formulation allows the model to learn the controlled modification of the existing image element while preserving the identity of the interior rather than generating any arbitrary new content.

However, turning off the artificial light is not a simple inverse transformation. When a light source is active, some information regarding shadows, local exposure, original material colors, and scene details may be distorted or lost. Therefore, in a computational context, turning off the light is less like an inverse style transfer and more akin to recovering an overexposed image, i.e., a task related to inpainting. This distinction motivates separate consideration of the tasks of turning the light on and off as related, but not entirely symmetrical, editing directions.

For this reason, our solution employs two distinct adapters: one responsible for turning on the artificial light and the other for turning it off. This separation prevents the mixing up of fundamentally different tasks within a single training signal and allows for more precise control over the editing outcome. At the same time, adapter specialization enables operation in a data-constrained setting, thereby reducing the volume and complexity of training data preparation.

Our work demonstrates that complex local light editing tasks can be addressed using relatively simple methods, provided the task is properly formulated and the approach is sufficiently decomposed. Rather than training a universal relighting model, we focus on a specific, practically relevant operation—changing the state of a particular artificial light source within an image—and show that even a small paired dataset can suffice to produce consistent and visually plausible results.

\section{Related Work}

\subsection{Image Editing and Relighting}
Image relighting is traditionally regarded as changing the scene illumination after capture. Unlike standard local edits, this operation requires coordinated changes in brightness, shadows, highlights, reflections, and color tone, while preserving the geometry, materials, and identity of the original scene. Interior images present a particular challenge: unlike outdoor scenes with a dominant light source, indoor spaces typically contain multiple local and global light sources, complex inter-reflections, soft shadows, and heterogeneous materials. This issue is discussed in detail in \textbf{ScribbleLight}~\cite{scribblelight2025}, where the authors emphasize that indoor relighting is complicated by a mixture of natural and artificial light, occlusions, and complex light interactions in cluttered scenes.

Early and physics-based approaches to relighting often rely on explicit modeling of geometry, materials, and lighting patterns. For instance, the \textbf{SwitchLight}~\cite{switchlight2024} portrait relighting method employs a physically grounded architecture based on Cook-Torrance reflection models, decomposing the image into intrinsic components for subsequent rendering under target lighting. While this approach offers physically interpretable control, it is specialized for portraits and requires explicit surface property descriptions, making it difficult to generalize to complex interior scenes containing numerous objects and materials.

Related research areas include harmonization and relighting techniques for image compositing. \textbf{Relightful Harmonization}~\cite{relightful2024} addresses the task of inserting a portrait subject into a new background by training a lighting-aware diffusion model; this model extracts lighting cues from the background image and uses them to harmonize the foreground lighting. The authors note that standard harmonization methods primarily adjust color, brightness, and contrast but often overlook lighting direction, intensity, and shadows, which can result in an unnatural-looking composite. Unlike such works, our task is not limited to harmonizing an object and a background; rather, it requires changing the state of light sources already present within a single scene.

\subsection{Diffusion Models for Lighting Control}

With the advancement of generative models, an increasing number of studies treat relighting as a conditional image editing task. \textbf{LightLab}~\cite{lightlab2025} proposes a diffusion-based method for the parametric control of light sources within a single image; the model is fine-tuned on a small set of real-world photo pairs—augmented by a significantly larger volume of synthetically rendered data—and leverages the linearity of light to synthesize controlled changes in light source intensity and color. This approach aligns with our problem formulation in that it addresses the modification of visible light sources and aims to provide explicit control over lighting parameters. However, LightLab relies on data augmentation via physically motivated synthetic rendering, whereas our work focuses on training a controllable LoRA adapter direction using a small paired dataset.

\textbf{ScribbleLight}~\cite{scribblelight2025} also employs a diffusion model for interior relighting but introduces user control via scribble annotations. The method combines albedo-conditioned Stable Diffusion with ControlNet, deriving the control signal from sketches and surface normals; this enables localized lighting adjustments, such as toggling lights and adding highlights, shadows, and indirect illumination. Unlike ScribbleLight, our method does not require spatial user masks or sketches during training; at the inference process, control is achieved by selecting the target object with a bounding box on the source image following the implementation of the base model.

\textbf{ControlLight}~\cite{controllight2026} treats lighting control as a task of controllable enhancement for low-light images. The method constructs a large-scale dataset \textbf{Light100K}: first, pseudo-paired normal-light targets are generated for real low-light images, and then intermediate states with varying lighting enhancement strengths are synthesized for each pair. For training, the authors employ FLUX.2-klein and fine-tune the diffusion model using LoRA; the enhancement strength parameter \(s\) serves as the LoRA adapter scale, enabling continuous adjustment of the image brightening level. Additionally, a misalignment-aware weighted flow matching loss is introduced to mitigate the impact of regions with boundary mismatches in the pseudo-target pairs and to help preserve the scene structure.
This approach resembles our work in that it also utilizes LoRA adaptation of a diffusion model for controllable lighting modification. However, ControlLight relies on a large pseudo-paired dataset and addresses the task of primarily global enhancement for low-light images by controlling the overall brightening strength.

\textbf{LuxRemix}~\cite{luxremix2026} proposes a method for interactive light editing in interior scenes by decomposing complex lighting into individual light source contributions—using a "One-Light-At-A-Time" (OLAT) format—and a background component. The method first trains a single-image lighting decomposition model, then propagates the resulting components across multiple scene views via multi-view lighting harmonization, and finally encodes the consistent lighting components into a relightable 3D Gaussian Splatting representation, allowing for interactive control over the intensity, color, and state of individual light sources. To train the single-image model, the authors use a large synthetic dataset comprising over 12,000 interior scenes featuring procedurally generated light sources and ground truth OLAT decompositions. The base model is a text-based image editing DiT fine-tuned using LoRA; the target source is specified via a spatial mask, while related tasks such as OLAT decomposition and turning off a selected source are distinguished by special text-based trigger words.
This method is the closest to our solution given the problem formulation; however, it still relies on large datasets, complex annotation, and modification of the base model, whereas we utilize a small, carefully curated dataset and lightweight adapters.

\subsection{Controllable Editing and LoRA Directions}

A distinct line of research focuses on identifying controllable directions within the parameter or latent space of diffusion models. \textbf{Concept Sliders}~\cite{conceptsliders2024} propose training LoRA adapters as interpretable "sliders" for precise control over visual attributes. Such a slider corresponds to a low-rank parameter direction that amplifies a chosen concept while minimizing the impact on other attributes, allowing for continuous and modular control. Our work is conceptually similar, as it treats LoRA not merely as a fine-tuning mechanism but also as a controllable editing direction. However, unlike Concept Sliders, we extract the LoRA editing direction directly within the model's flow/velocity field, linking two opposing scene states: lights on and lights off.

\textbf{LoRA of Change}~\cite{loraofchange2026} is the work closest in spirit to ours. The authors view LoRA as a carrier of visual instructions derived from before-after pairs. Their method dynamically generates an instruction-specific LoRA from a single before-after pair and introduces LoRA Reverse Training to learn from paired data without requiring more complex quad sets. In contrast, however, we do not generate a LoRA for an arbitrary instruction using a hypernetwork; instead, we train a compact LoRA for a specific control operation and explicitly extract a signed direction in velocity/flow space via symmetric +s/-s passes. This enables the operation in a tiny-data regime and allows the sign of the adapter weight to serve as a control variable for toggling the scene state.

\subsection{Flow Matching and FLUX.1 Kontext}

We use \textbf{FLUX.1 Kontext}~\cite{fluxkontext2025} as our base model — a flow-matching~\cite{lipman2023flowmatching} model designed for in-context image generation and editing within latent space. FLUX.1 Kontext integrates image generation and editing into a unified architecture that leverages both textual and visual context; it demonstrates strong performance in local and global editing, object identity preservation, and stability across multi-step edits. These characteristics make it a suitable foundation for interior editing tasks, which require altering the lighting while preserving the composition, geometry, and objects of the scene.

Our work advances this line of research by treating the LoRA~\cite{hu2021lora} adapter as a directional addition to the flow/velocity field of the base model. Unlike the standard application of LoRA as a static fine-tuning mechanism, we interpret the sign and scale as control parameters for editing of the adapter. A symmetric procedure involving +s and -s passes allows us to isolate the pure direction of the LoRA effect without relying on the zero-adapter state as a teacher. This is particularly crucial for lighting modification tasks, where the base model (without the adapter) does not necessarily define a correct neutral trajectory between the lights on and lights off states.

\paragraph{Positioning Our Method}

Thus, the existing methods for image lighting control can be broadly categorized into three groups: physically motivated relighting approaches, diffusion-based methods utilizing external control signals, and controlled editing methods via LoRA directions. Physically motivated methods offer interpretability but often require specialized setups, additional scene representations, or complex data. Diffusion-based relighting methods provide a stronger photorealistic prior but typically rely on explicit masks, sketches, normal maps, large datasets, or parametric light source descriptions. LoRA-based methods enable the compact encoding of visual transformations; however, in most cases, they do not consider the problem formulation itself as a distinct factor determining the quality and robustness of the solution.

Our method occupies an intermediate position between these directions. On the one hand, we use the photorealistic prior of a modern diffusion model and do not build an explicit physical model of the scene. On the other hand, we do not reduce the task to universal text editing, but clearly decompose it into two semantically related, but different in form transformations: turning on artificial light - changing the style, and turning it off - restoring the overexposed image. We focus on the correct formulation of the problem and its formalization to determine ways to implement the solution.
We use Flux.1 Kontext as the base model, relying on its excellent ability to preserve details of the original scene during local image editing, and train two adapters with separate areas of application on its basis.
For the task of turning on the light, we consider LoRA not just as a set of additional weights, but as a directional correction to the flow/velocity field of the base diffusion model. Unlike approaches where the adapter effect is assessed by comparing a model with and without LoRA, we perform two symmetric runs with positive and negative adapter scales. The LoRA edit direction is then extracted as the central difference between the corresponding flow/velocity fields. This avoids using the zero-adapter state as a teacher, which is especially important in cases where the base model does not specify the correct neutral direction for the target operation.
The resulting direction is interpreted as a controllable addition to the flow field of the base model, responsible for transitioning the selected light source to the active state. To improve direction stability, we use an A/B training scheme that connects forward and reverse examples of lighting changes, as well as the loss function that combines supervised flow-matching loss and direction-tying loss. In this case, the negative adapter scale is used not as an independent solution to the task of turning off the light, but as a symmetric means of stabilizing and extracting the direction for the light-on operation.
For the task of turning off the light, a separate LoRA adapter and a separate training set-up are used. It is based on the standard inpainting-like pipeline, but is complemented by a loss function specifically adapted to artificial light editing. This function enhances the contribution of areas associated with the light source and its effect on the scene, but does not completely nullify the background, so that the model continues to be signaled to preserve the geometry, materials and composition of the frame. Additionally, low-frequency matching of the predicted and target flow fields is used, which penalizes errors in global exposure, color temperature and shadow distribution. Thanks to this, the adapter learns not just to darken the lamp area, but to consistently restore the visual state of the scene after removing the contribution of the artificial source.
The key difference of our method lies not in making the architecture more complex, but in combining three simple solutions: using a strong base image-editing model, training compact LoRA adapters on a small paired dataset, and formulating the light-on and light-off tasks separately. This approach enables local editing of artificial lighting without large composite datasets, complex physical scene reconstruction, dense masks, sketch annotations, or multi-view capture, while preserving the controllability of the result and the identity of the original interior.
In view of the above, it is important to emphasize that we use the control bounding box only at the inference stage, and it is not part of the LoRA adapter training signal. During training, the target light source is specified exclusively by a text instruction, and the source images are not supplemented with masks, boxes, or other spatial annotations. At the inference stage, the selected luminaire is spatially localized using the existing capabilities of the base FLUX.1 Kontext model. Thus, the adapter learns the direction of the lighting-state change itself, while the mechanism for selecting a specific object remains with the base image-editing model.

\section{Dataset}
In this work, we were limited in our ability to prepare and utilize a large-scale training dataset. Consequently, we sought to compensate for the limited sample size by prioritizing the quality and controllability of the constituent examples. This resulted in the creation of a paired dataset comprising approximately 320 image pairs. The majority (around $90\%$) of the dataset consisted of specially prepared synthetic examples, while the remaining $10\%$ comprised real-world photographs. This approach allowed us to augment the controlled synthetic samples with examples featuring natural light and shadow distributions.

Each image pair depicts the same scene under two different artificial lighting conditions and is used to construct corresponding forward and reverse training examples. This format enables control over the state of a selected light source while preserving the geometry, materials, and overall composition of the scene.

It is important to note that the training dataset consisted exclusively of interior scenes. Thus, at the current stage of the method development, its intended application is limited to editing artificial light sources located indoors.
Nevertheless, a number of exterior scenes were deliberately included in the final test set to better evaluate the generalization limits of the method. See examples of the method performance in Appendices A and B.

\subsection{Synthetic Data}

The synthetic subset was generated using specially selected, high-resolution, detailed interior scenes processed to closely approximate the visual characteristics of real-world photographs.

To increase the diversity of lighting sources, the synthetic portion of the dataset was divided into several categories corresponding to the most common types of interior light fixtures found in residential and public spaces:

\begin{itemize}
\item Table lamps (\textit{table});
\item Ceiling lights (\textit{ceiling});
\item Floor lamps (\textit{floor});
\item Linear interior lights (\textit{linear});
\item Pendant lights (\textit{pendant});
\item Wall lights (\textit{wall}).
\end{itemize}

However, the dataset does not include light sources that are specialized or atypical for the interior layouts under consideration, such as decorative multi-lamp chandeliers, candle-shaped bulbs, or industrial and medical fluorescent fixtures.

Each category comprises scenes featuring the respective light sources under varying environmental conditions and focal lengths, and with varying degrees of contribution to the overall illumination of the room. Consequently, the dataset encompasses both local light sources which primarily affect nearby surfaces and more prominent sources that significantly alter the global lighting balance of the scene.

This categorization aims to increase the diversity of geometry and artificial light propagation patterns within the unified training dataset.

\subsection{Real world data}

Real-world image pairs of lamps and light fixtures captured in natural settings account for approximately $10\%$ of the training dataset. Unlike the synthetic component, generating this data requires capturing the same scene under varying artificial lighting conditions. Moreover, when working with very small datasets, alignment errors such as those caused by a camera shake during capture can have catastrophic consequences for the final adapter. This makes collecting real-world pairs significantly more labor-intensive than collecting and annotating individual images. Consequently, this portion of the dataset is limited in both the total number of images and the diversity of light source types represented.

Despite their smaller volume and less controlled capture conditions, real-world images play a crucial role in training. Since light is a complex physical phenomenon its modeling in synthetic scenes inevitably entails limitations regarding the accurate reproduction of light transport. Even with highly realistic rendering, the lighting distribution in synthetic data may remain more regular and predictable than in real-world photographs. In contrast, real-world images directly reflect the actual behavior of light within a scene, including the interaction of multiple sources, reflections from various materials, shadow non-uniformity, local exposure variations, and other effects that are difficult to fully replicate through synthetic generation.

Thus, the synthetic and real-world components of the dataset serve complementary functions. Synthetic data provides precise control over paired scene states and increases the diversity of training examples, while real-world photographs help bridge the gap between synthetic and real-world lighting distributions.

\section {Light-On}
\subsection{Task Formulation}

Let's consider the task of controllable image editing that involves turning on artificial light sources in a scene. Let a set of paired examples be given.

\[
\mathcal{D}=\{(x_i^{0}, x_i^{1}, p_i^{+}, p_i^{-})\}_{i=1}^{N},
\]
where \(x_i^{0}\) denotes the initial scene image without the target lighting effect or with the light source turned off, and \(x_i^{1}\) denotes an image of the same scene after the artificial lighting is turned on. The text conditions \(p_i^{+}\) and \(p_i^{-}\) describe, respectively, the forward operation of turning on the light and the reverse operation of restoring the initial state.

We consider a conditional diffusion model for image editing in which the context image \(c\) defines the preserved scene structure, while the target image \(x\) defines the editing result. Pairs used for forward task \(A\)
\[
c_A=x^0,\qquad x_A=x^1,\qquad p_A=p^+,
\]
and for inverse task \(B\):
\[
c_B=x^1,\qquad x_B=x^0,\qquad p_B=p^-.
\]

A distinctive feature of the task at hand is that turning on artificial light is not merely a matter of locally altering the color or brightness of a specific object. It necessitates a coordinated shift in the illumination distribution of the scene involving the appearance of a light spot, the over-illumination of nearby surfaces, shadows, reflections, color shifts, and contrast changes. In practice, a baseline model often lacks a robust representation of this operation: it might add a visible light source without physically consistent effects on the surroundings, modify textures rather than lighting, or interpret the request to turn on the light as a standard object-level edit.

The objective is to train a LoRA adapter with parameter \(\phi\) that defines a controllable editing direction. A positive adapter multiplier corresponds to the turn on light operation, whereas a negative multiplier is employed solely during training to evaluate and stabilize the LoRA-induced direction. We do not treat the negative scaling of this adapter as the primary mechanism for turning the light off. Consequently, beyond the standard supervised editing loss, a key element of the formulation is the requirement for directional consistency in the LoRA-induced change. We formulate this requirement as \emph{directional consistency regularization}: the direction of change that the LoRA introduces into the model velocity field for the forward operation must be consistent with the direction for the reverse operation. For the task of turning lights on and off, this consistency is antisymmetric in nature.

\subsection{Method}

Let \(\mathcal{E}\) denote the VAE encoder that maps an image \(x\) to a latent representation.
\[
z=\mathcal{E}(x).
\]
During the training step, a timestep \(t\) and noise \(\epsilon\) are selected. The scheduler determines the noisy latent.
\[
z_t = q_t(z,\epsilon),
\]
where the specific form of \(q_t\) is determined by the flow-matching or diffusion parameterization employed.

The {Flux Kontext}~\cite{fluxkontext2025} model takes a noisy latent, a timestep, a text condition, and a context image as input. Let us denote its prediction as
\[
v_{\theta,\phi}(z_t,t,p,c;a),
\]
where \(\theta\) denotes the frozen parameters of the base model, \(\phi\) denotes the trainable LoRA parameters, and \(a\) denotes the scalar multiplier of the adapter.

The key idea of the method is to explicitly isolate the direction that LoRA adds to the velocity field predicted by the model. To do this, we use the symmetric difference between two predictions computed on the same input but with opposite adapter multipliers. For a fixed \(s>0\), we define the LoRA-induced direction:
\[
\Delta_{\phi}(z_t,t,p,c)
=
\frac{
v_{\theta,\phi}(z_t,t,p,c;+s)
-
v_{\theta,\phi}(z_t,t,p,c;-s)
}{2s}.
\]
Both predictions in the numerator are computed using the same \(z_t\), \(t\), \(\epsilon\), \(p\), and \(c\); only the sign of the adapter multiplier differs. Such a central difference suppresses the common component of the base model and estimates the local LoRA-induced direction in the model prediction space.

For forward task \(A\) and inverse task \(B\), we obtain:
\[
\Delta_A =
\frac{v_A(+s)-v_A(-s)}{2s},
\qquad
\Delta_B =
\frac{v_B(+s)-v_B(-s)}{2s}.
\]
Since the forward and backward passes are computed for a single training pair with consistent values of \(t\) and \(\epsilon\), \(\Delta_A\) and \(\Delta_B\) can be viewed as coupled estimates of the LoRA-induced response for two opposing states of the same scene.
\[
\mathcal{L}_{\mathrm{dir}}
=
\left\|
\Delta_A+\Delta_B
\right\|_2^2.
\]

The concept of using forward and inverse transformations as mutually consistent training signals traces back to earlier works on cycle consistency, specifically CycleGAN~\cite{zhu2017cyclegan}. In our work, this principle is not applied directly but is instead reinterpreted for modern flow-matching image-editing models and LoRA parameterization. Unlike classical cycle consistency, we do not construct an explicit closed loop where the output of the forward transformation serves as the input for the inverse transformation to reconstruct the original image. Instead, the forward and inverse tasks are trained on the corresponding ground-truth pairs, and their mutual consistency is enforced at the level of LoRA-induced directions within the flow/velocity field of the model. Thus, rather than requiring $F(G(x)) \approx x$, we employ an antisymmetric constraint—$\Delta_A \approx -\Delta_B$, which is more natural for our formulation; this constraint links the model's responses for opposite directions of the scene state change.

We utilize this coupling as a regularizer. Its role is to encourage opposing response trends in the model prediction space for the forward and reverse operations, thereby establishing a more robust editing direction associated with changes in the light source state. Without such a constraint, the supervised loss could be satisfied through localized color or texture changes without forming a coherent representation of the target operation.

In addition to directional consistency regularization, we employ the standard supervised flow-matching loss~\cite{lipman2023flowmatching}. For the forward task:
\[
\mathcal{L}_A
=
\ell\left(v_A(+s), u_A\right),
\]
where \(u_A\) is the target velocity/flow value determined by the scheduler. For reverse pairs, an additional auxiliary supervised term may be used:
\[
\mathcal{L}_B
=
\ell\left(v_B(-s), u_B\right).
\]
The final loss function has the form
\[
\mathcal{L}
=
w_A\mathcal{L}_A
+
w_B\mathcal{L}_B
+
w_{\mathrm{dir}}\mathcal{L}_{\mathrm{dir}},
\]
where \(w_A\), \(w_B\), and \(w_{\mathrm{dir}}\) specify the weights for the forward supervised loss, the backward supervised loss, and the directional consistency regularization, respectively.

Crucially, the method does not use the base model prediction—with LoRA disabled—as a teacher. In tasks involving turning on artificial light, such a prediction often serves as an unreliable reference point, as the base model may not grasp the act of turning on the light as a physically consistent redistribution of scene illumination. Instead, we compare two symmetric states of the same LoRA, \(+s\) and \(-s\); this allows us to learn a directed and controllable change in the velocity field rather than an absolute correction relative to relatively weak behavior of the base model.
It is important to emphasize that, in this formulation, inverse pairs serve as an auxiliary regularization signal that stabilizes the learned editing direction. At the inference stage, turning off the light is not achieved by applying a negative scale to the Light-On adapter; instead, a separate Light-Off adapter with its own formulation and loss function is trained for that operation.

\section {Light-Off}
\subsection{Task Formulation}

We consider the task of conditional image editing, which involves turning off an artificial light source while preserving the scene geometry, object materials, and frame composition. Let \(x_i^{\mathrm{on}}\) denote the original image of the scene with the light source on, and \(x_i^{\mathrm{off}}\) denote the target image of the same scene after the light source is turned off. The training set is defined by a collection of pairs
\[
\mathcal{D} =
\left\{
\left(x_i^{\mathrm{on}}, x_i^{\mathrm{off}}, p_i, m_i\right)
\right\}_{i=1}^{N},
\]
where \(p_i\) is a text editing instruction, and \(m_i\) is an optional spatial loss weight map.

The goal is to learn the parameters of adapter \(\phi\) which, when added to a fixed pre-trained conditional editing model \(F_{\theta_0}\), implements the transformation
\[
T_{\phi}: x^{\mathrm{on}} \mapsto x^{\mathrm{off}}.
\]
In this process, the parameters of the base model \(\theta_0\) remain unchanged, and only the low-rank LoRA increment is trained.
\[
\theta(\phi) = \theta_0 + \Delta \theta_{\phi}.
\]

Unlike localized tasks such as object removal or replacement, turning off a light source requires altering the global lighting, specifically brightness, color temperature, shadows, reflected light, and the glow of the source itself. Consequently, the task cannot be reduced to simply darkening the area around the lamp. Furthermore, the scene identity, including object positions and shapes, materials, textures, and overall composition, must be preserved.

It is also important to note that the inverse transformation (\(x^{\mathrm{off}} \mapsto x^{\mathrm{on}}\)) is not strictly reversible in the general case. A single image with the light source turned off could correspond to various lighting configurations, intensities, and reflection conditions. Therefore, the task is primarily framed as a unidirectional editing task (\(x^{\mathrm{on}} \rightarrow x^{\mathrm{off}}\)) rather than the learning of a symmetric pair of mutually inverse transformations.

\subsection{Method}

We use a pre-trained model for conditional image editing in latent space. Let \(E\) denote the VAE encoder that maps an image to a latent representation. For each training pair, the target latent is defined as
\[
z_i = E\left(x_i^{\mathrm{off}}\right),
\]
and the image \(x_i^{\mathrm{on}}\) is used as an editing condition:
\[
c_i = x_i^{\mathrm{on}}.
\]

At each training step, a timestep \(t\) and noise are selected.
\[
\epsilon \sim \mathcal{N}(0, I).
\]
The noisy latent is constructed in the standard manner for diffusion or flow-matching training~\cite{lipman2023flowmatching}:
\[
z_t = \alpha_t z_i + \sigma_t \epsilon,
\]
where \(\alpha_t\) and \(\sigma_t\) are determined by the scheduler used.

The model with the LoRA adapter predicts the training signal.
\[
\hat{v}_{\phi}
=
F_{\theta_0 + \Delta \theta_{\phi}}
\left(
z_t, t, p_i, c_i
\right).
\]
Let us denote the target signal by \(v_i(t)\). Depending on the parameterization of the base model, this may be noise, velocity, or another quantity used in the standard loss function of the model.

The primary supervised loss function has the form
\[
\mathcal{L}_{\mathrm{sup}}
=
\mathbb{E}_{i,t,\epsilon}
\left[
\left\|
w_i \odot
\left(
\hat{v}_{\phi} - v_i(t)
\right)
\right\|_2^2
\right],
\]
where \(w_i\) is a spatial weight map derived from \(m_i\) (or set to unity in the absence of a mask), and \(\odot\) denotes element-wise multiplication. The mask is normalized such that its mean value is close to one. This allows for emphasizing regions associated with the light source and its influence without completely excluding the background from the training process.

For the task of realistically turning off the light, a single local supervised loss in the training signal space is insufficient. While such a loss can ensure correspondence between the predicted and target flow/velocity signals at individual spatial locations, it does not guarantee the coherent reconstruction of global scene lighting—such as exposure, color temperature, soft shadows, and reflected light. Therefore, low-frequency and statistical loss components are introduced to encourage the alignment of larger-scale lighting structures between the prediction and the target state.
\[
\mathcal{L}_{\mathrm{lf}}
=
\lambda_{\mathrm{lf}}
\sum_{k \in \mathcal{K}}
\rho
\left(
P_k\left(\hat{v}_{\phi}\right),
P_k\left(v_i(t)\right)
\right),
\]
where \(P_k\) is average pooling with a window size of \(k\), \(\mathcal{K}\) is a set of scales (e.g., \(\{8,16,32\}\)), and \(\rho\) is the \(L_1\) or \(L_2\) distance.

Additionally, global channel-wise statistics are matched:
\[
\mathcal{L}_{\mathrm{stat}}
=
\lambda_{\mathrm{stat}}
\left(
\left\|
\mu\left(\hat{v}_{\phi}\right) - \mu\left(v_i(t)\right)
\right\|_1
+
\left\|
\sigma\left(\hat{v}_{\phi}\right) - \sigma\left(v_i(t)\right)
\right\|_1
\right),
\]
where \(\mu(\cdot)\) and \(\sigma(\cdot)\) denote the mean and standard deviation across spatial coordinates. This term helps stabilize the global exposure and color temperature of the image.

To prevent the removal or distortion of the light source object itself, identity-preservation augmentation is employed. With probability \(q\), the training pair is replaced by an identity-editing pair:
\[
\left(x_i^{\mathrm{on}}, x_i^{\mathrm{off}}, p_i\right)
\rightarrow
\left(x_i^{\mathrm{on}}, x_i^{\mathrm{on}}, p_{\varnothing}\right),
\]
where \(p_{\varnothing}\) is a neutral or empty instruction. In this case, the target latent becomes
\[
z_i^{\mathrm{id}} = E\left(x_i^{\mathrm{on}}\right).
\]
This step trains the adapter to adopt the permissible "make no changes" mode and reduces the likelihood of the model removing the lamp as an object instead of simply turning off its emission.

The final loss function is written as
\[
\mathcal{L}
=
\mathcal{L}_{\mathrm{sup}}
+
\mathcal{L}_{\mathrm{lf}}
+
\mathcal{L}_{\mathrm{stat}}
+
\mathcal{L}_{\mathrm{id}},
\]
where \(\mathcal{L}_{\mathrm{id}}\) has the same form as \(\mathcal{L}_{\mathrm{sup}}\) but is computed on identity-augmentation steps.

\section{Metrics and Evaluation Protocol}
\label{sec:light_edit_evaluation}

Since LightLab~\cite{lightlab2025} and LuxRemix~\cite{luxremix2026} do not publicly release implementations of their methods or complete test datasets containing ground truth (the absolute standard of correctness), the comparison was conducted using images available in their published test examples. We independently tested our own method, as well as available open-source models, on those images. Thus, the scope of the evaluation was determined by the actual availability of the source images and the results produced by the proprietary solutions.

As there were no reference images with correctly modified lighting conditions available for the task at hand, standard full-reference metrics (such as PSNR and SSIM~\cite{wang2004ssim}) could not be used. The evaluation combined three complementary groups of metrics: no-reference visual quality metrics, vision-language metrics for instruction adherence, and a structured assessment by a multimodal language model. Spatial masks representing light sources, lighting propagation areas, or static backgrounds were not used in calculating the reported results. This protocol allows for the separate assessment of visual quality of the result, the accuracy of the requested lighting modification, and the plausibility of the generated image.

Three no-reference IQA metrics were used to evaluate overall visual quality: MUSIQ~\cite{ke2021musiq}, ARNIQA~\cite{agnolucci2024arniqa}, and LIQE~\cite{zhang2023liqe}. MUSIQ analyzes the image at multiple spatial scales using a transformer architecture and assesses quality by considering both local distortions and the overall image structure. ARNIQA employs a self-learned distortion space representation and characterizes the presence of unnatural degradations independently of scene content. LIQE evaluates quality based on the alignment of visual and textual representations, simultaneously accounting for the scene type and the nature of the distortions. For all three metrics, a higher value corresponds to higher perceived quality. Notably, these metrics are task-agnostic: a visually high-quality image can receive a high score even if the light source state is altered incorrectly.

Instruction alignment was evaluated using Directional CLIP and AugCLIP. Directional CLIP utilizes the CLIP representation space~\cite{radford2021clip} and computes the cosine similarity between the direction of the image change and the direction of the requested text-based transition~\cite{gal2022stylegannada}:

\[
S_{\mathrm{DirCLIP}} =
\cos\left(
f_I(I_{\mathrm{edit}})-f_I(I_{\mathrm{src}}),
f_T(T_{\mathrm{target}})-f_T(T_{\mathrm{source}})
\right),
\]

where $f_I$ and $f_T$ denote the CLIP visual and text encoders,
$I_{\mathrm{src}}$ and $I_{\mathrm{edit}}$ are the source and
edited images, and $T_{\mathrm{source}}$ and
$T_{\mathrm{target}}$ are the descriptions of the source and target
lighting states. Unlike standard CLIP similarity, this metric evaluates
the direction of the change made, rather than just the properties of the
final image. However, the use of global representations limits its
sensitivity to the locality of the edit.

AugCLIP~\cite{kim2025augclip} constructs a context-aware
representation of the intended ideal result by combining features from
the source image and the target state. The post-edit image is compared
against this representation within the CLIP space. Thus, the metric
accounts for both instruction adherence and the preservation of the
source scene's content. A single representation of the target state was
used for the same "source image–instruction" pair when evaluating all
compared methods.

Additionally, the original and edited images were evaluated by the multimodal model \texttt{gpt-5.5} via the OpenAI Responses API. The model generated structured scores based on four criteria. The success of switching the required light source (\textit{State success}) was rated on a scale of 0 to 2. Physical lighting plausibility (\textit{Physical consistency}), the preservation of scene geometry, objects, materials, and details (\textit{Scene preservation}), and the photorealism of the result (\textit{Photorealism}) were rated on a scale of 1 to 5. A binary flag, \textit{Global exposure cheat}, was also recorded; this indicated instances where the required local change was predominantly replaced by a global brightening, darkening, or color shift of the entire scene. The model was provided with a precise instruction, the original image, and the edited result, but did not receive spatial masks.

The raw metrics have different ranges; therefore, prior to aggregation, they were normalized to a fixed $[0,1]$ scale, where a higher value indicates a better result:

\[
\begin{aligned}
\widetilde{S}_{\mathrm{MUSIQ}}
    &= \frac{S_{\mathrm{MUSIQ}}}{100},
&\qquad
\widetilde{S}_{\mathrm{ARNIQA}}
    &= S_{\mathrm{ARNIQA}}, \\[2pt]
\widetilde{S}_{\mathrm{LIQE}}
    &= \frac{S_{\mathrm{LIQE}}-1}{4},
&
\widetilde{S}_{\mathrm{DirCLIP}}
    &= \frac{S_{\mathrm{DirCLIP}}+1}{2}, \\[2pt]
\widetilde{S}_{\mathrm{AugCLIP}}
    &= \frac{S_{\mathrm{AugCLIP}}+1}{2},
&
\widetilde{S}_{\mathrm{State}}
    &= \frac{S_{\mathrm{State}}}{2}, \\[2pt]
\widetilde{S}_{k}
    &= \frac{S_k-1}{4},
&\quad
k &\in
\{\mathrm{Physics},\mathrm{Preservation},\mathrm{Photo}\}, \\[2pt]
S_{\mathrm{Compliance}}
    &= 1-p_{\mathrm{cheat}}.
\end{aligned}
\]

where $p_{\mathrm{cheat}}$ is the proportion of images for which a global exposure shift was detected. Fixed normalization was used rather than min–max normalization based on the sample under consideration. Thus, the value for a given model does not directly depend on the results of the other models.

The normalized values were first aggregated within semantic groups. The integrated IQA score was the average of the MUSIQ, ARNIQA, and LIQE values, while the VLM score was the average of four structured visual analysis criteria:

\begin{align*}
S_{\mathrm{IQA}} &=
\frac{
\widetilde{S}_{\mathrm{MUSIQ}}+
\widetilde{S}_{\mathrm{ARNIQA}}+
\widetilde{S}_{\mathrm{LIQE}}
}{3}, \\
S_{\mathrm{VLM}} &=
\frac{
\widetilde{S}_{\mathrm{State}}+
\widetilde{S}_{\mathrm{Physics}}+
\widetilde{S}_{\mathrm{Preservation}}+
\widetilde{S}_{\mathrm{Photo}}
}{4}.
\end{align*}

The final indicator was calculated as a weighted sum of the group assessments:

\[
S_{\mathrm{total}} =
0.40S_{\mathrm{VLM}}+
0.20S_{\mathrm{IQA}}+
0.20\widetilde{S}_{\mathrm{AugCLIP}}+
0.10\widetilde{S}_{\mathrm{DirCLIP}}+
0.10S_{\mathrm{Compliance}}.
\]

The highest weight was assigned to the VLM component, as it directly
evaluates the execution of the target edit and the physical correctness
of the lighting. IQA and AugCLIP were each assigned a weight of $0.20$:
the former characterizes technical and perceptual image quality,
while the latter reflects the balance between instruction adherence
and the preservation of original content. Directional CLIP was used
to additionally assess the direction of change, with a weight of $0.10$.
Another $0.10$ was allocated to adherence to the constraint against
global exposure changes. These coefficients are heuristic and reflect
the priorities of the task at hand, rather than being statistically
learned or universally optimal weights.

Tables~\ref{tab:light_on} and~\ref{tab:light_off} present the average
raw metric values prior to normalization and the final normalized
score $S_{\mathrm{total}}$. Arrows indicate the preferred direction
of change for each metric.

\begin{table*}[!t]
\centering
\caption{Quantitative assessment of the results of turning on the light.
The initial metrics are presented prior to normalization.}
\label{tab:light_on}
\scriptsize
\setlength{\tabcolsep}{3.2pt}
\resizebox{\textwidth}{!}{%
\begin{tabular}{l*{11}{r}}
\toprule
Method &
MUSIQ $\uparrow$ &
ARNIQA $\uparrow$ &
LIQE $\uparrow$ &
DirCLIP $\uparrow$ &
AugCLIP $\uparrow$ &
State $\uparrow$ &
Physics $\uparrow$ &
Preserv. $\uparrow$ &
Photo $\uparrow$ &
Cheat $\downarrow$ &
$S_{\mathrm{total}}\uparrow$ \\
\midrule
\textbf{Ours}
& 63.649 & 0.653 & 3.976 & 0.0355 & \textbf{0.7986}
& 1.923 & \textbf{1.692} & \textbf{1.692} & \textbf{1.692} & \textbf{7.7\%} & \textbf{0.6076} \\
LightLab~\cite{lightlab2025}
& \textbf{71.370} & \textbf{0.698} & \textbf{4.481} & 0.0329 & 0.7937
& \textbf{2.000} & 1.400 & 1.500 & 1.600 & 30.0\% & 0.5907 \\
Qwen
& 68.247 & 0.659 & 3.765 & \textbf{0.0393} & 0.7826
& 1.833 & 1.417 & 1.333 & 1.667 & 16.7\% & 0.5762 \\
Flux2Klein
& 69.332 & 0.663 & 4.260 & 0.0362 & 0.7621
& 1.700 & 1.400 & 1.500 & 1.600 & 20.0\% & 0.5753 \\
LuxRemix~\cite{luxremix2026}
& 63.283 & 0.670 & 3.596 & 0.0310 & 0.7776
& 1.750 & 1.000 & 1.000 & 1.000 & 50.0\% & 0.4969 \\
\bottomrule
\end{tabular}%
}
\end{table*}

\begin{table*}[!t]
\centering
\caption{Quantitative assessment of the results of turning off the lights.
The initial metrics are presented prior to normalization.}
\label{tab:light_off}
\scriptsize
\setlength{\tabcolsep}{3.2pt}
\resizebox{\textwidth}{!}{%
\begin{tabular}{l*{11}{r}}
\toprule
Method &
MUSIQ $\uparrow$ &
ARNIQA $\uparrow$ &
LIQE $\uparrow$ &
DirCLIP $\uparrow$ &
AugCLIP $\uparrow$ &
State $\uparrow$ &
Physics $\uparrow$ &
Preserv. $\uparrow$ &
Photo $\uparrow$ &
Cheat $\downarrow$ &
$S_{\mathrm{total}}\uparrow$ \\
\midrule
\textbf{Ours}
& 66.716 & 0.678 & 4.121 & 0.0189 & 0.6509
& 1.688 & \textbf{1.188} & 1.188 & 1.438 & \textbf{31.25\%} & \textbf{0.5312} \\
LightLab~\cite{lightlab2025}
& 63.350 & 0.660 & 3.470 & 0.0488 & 0.6111
& 1.667 & 1.000 & \textbf{1.333} & \textbf{1.667} & 31.3\% & 0.5179 \\
Flux2Klein
& \textbf{70.455} & 0.673 & \textbf{4.681} & 0.0263 & \textbf{0.6624}
& 1.688 & 1.000 & 1.063 & 1.313 & 62.5\% & 0.5020 \\
LuxRemix~\cite{luxremix2026}
& 47.042 & \textbf{0.696} & 3.340 & 0.0481 & 0.4002
& \textbf{2.000} & 1.000 & 1.000 & 1.000 & 35.0\% & 0.4742 \\
Qwen
& 53.304 & 0.557 & 2.387 & \textbf{0.0693} & 0.6211
& 1.500 & 1.000 & 1.000 & 1.125 & 93.8\% & 0.3957 \\
\bottomrule
\end{tabular}%
}
\end{table*}

\FloatBarrier
Comparative image examples for both tasks are provided in Appendices A and B.

\section{Limitations}

The primary limitation of the proposed approach stems from the use of LoRA adapters trained on a dataset that is relatively small and limited in diversity. Although this training format yields a compact and controllable editing direction, the quality of the result depends significantly on how well the target light source type is represented in the training set.

Specifically, the adapter generalizes poorly to light fixtures with geometries and emission characteristics that differ substantially from the examples seen during training. As shown in Appendix A, this is particularly evident with non-standard sources—such as chandeliers with candle-style bulbs or long wall-mounted fluorescent lights—that were entirely absent from the training data. In such cases, the model may be less precise in altering the state of the light source itself or may fail to generate a sufficiently coherent light distribution within the surrounding scene.

This limitation is an expected consequence of narrow specialization of the LoRA adapter: while it effectively encodes transformations characteristic of the training data distribution, it lacks robustness when applied to significantly different types of light sources. Consequently, the current method should not be viewed as a fully universal solution for arbitrary artificial lighting configurations.

The most direct way to mitigate this limitation is to expand the training dataset by incorporating a greater variety of luminaire types, lampshade shapes, light source geometries, and placement configurations within the scene. This also aligns with the direction of future work focused on increasing the volume and diversity of the data.

\section{Ablation Study}

To evaluate the contribution of individual components of the proposed method, we conducted a series of ablation experiments. The primary goal was to determine which elements of the training scheme were essential for establishing a stable editing direction and preserving lighting consistency, and which modifications yielded no noticeable improvement in the results.

\subsection{Using Base Model as Reference State}

One of the approaches we considered involved using the model prediction with a zero LoRA adapter scale as the reference state for estimating the direction of change. This formulation aligns with a more straightforward interpretation of LoRA as a deviation from the original model behavior.

The experiments revealed that this approach resulted in lower quality compared to the base model. The reason is that the state with a zero adapter does not serve as a reliable neutral point for the operation in question. The base model does not always correctly conceptualize the activation of an artificial light source as a coherent change in the scene state; instead, it may interpret the instruction as a localized alteration in brightness, texture, or the appearance of the light fixture itself.

Consequently, using a zero adapter introduces an erroneous reference component into the learned direction. This observation motivated a shift toward a symmetric scheme that calculates the central difference between forward passes using positive and negative adapter scales.

\subsection{Combining Forward and Inverse Tasks in Single Adapter}

We also investigated the feasibility of combining the light-activation and light-deactivation operations within a single LoRA adapter controlling the transformation direction solely through the sign of the adapter scale. In this setup, a positive adapter coefficient was used to turn the light on, while a negative coefficient was used to turn it off.

For the light-on operation, this variant did not result in noticeable degradation and maintained comparable quality. However, for the light-off operation, quality dropped significantly. This confirms that the two operations are not symmetrical in a machine learning context.

Turning on the light corresponds more closely to a directed change in the visual state of an existing scene, whereas turning it off requires recovering information that has been partially obscured or distorted by the active light source. Specifically, it is necessary to reconstruct local exposure, shadows, material colors, and reflected lighting. Consequently, using a single adapter with LoRA sign control proves insufficient for the effective inverse task solution.

This experiment confirms the advisability of training separate Light-On and Light-Off adapters.

\subsection{Normalization Frequency}

We also investigated the impact of applying normalization more frequently during the training process. It was hypothesized that increasing the normalization frequency might stabilize the magnitude of the learned direction and reduce variance between updates.

In practice, more frequent normalization did not lead to any noticeable improvement in visual quality or training stability. Moreover, it increased computational costs and slowed down training. Thus, the additional normalization proved unnecessary; the final configuration employed a less frequent normalization scheme that delivered comparable quality at a lower computational cost.

\subsection{Contribution of Reverse-Scheme Components}

To verify the necessity of the individual components of the reverse scheme, we sequentially removed the additional elements employed, including the regularizing loss functions and direction normalization.

Removing these elements consistently led to a degradation in editing quality. In the limiting case, the training procedure effectively reduced to standard LoRA fine-tuning on image pairs. While this approach could learn superficial statistical differences between states, it failed to establish a stable direction corresponding to a physically and visually consistent change in artificial lighting.

In practice, this manifested as localized changes in the light source brightness or color without correctly altering surrounding surfaces, shadows, or reflections. Thus, the observed degradation demonstrates that the additional reverse-training structure is essential not only for stabilizing optimization but also for preventing the task from degenerating into standard LoRA fine-tuning.

\subsection{Impact of Real-World Images in Training Buffer}

To assess the impact of real-world data, we conducted a separate series of experiments in which real-world photographs were excluded from the training buffer. Training exclusively on synthetic pairs resulted in poorer model generalization to real-world images.

Specifically, the model was less effective at reproducing light propagation patterns and shadow formation in real scenes. Despite the high visual realism of the synthetic data, the lighting distribution within them remained more regular and predictable. Incorporating even a relatively small number of real examples bridged this gap and improved the consistency between the light source state and its effect on surrounding surfaces.

The obtained result aligns with the observations presented in ControlLight~\cite{controllight2026}: while synthetic data enable the scaling of the training set and the generation of controllable pairs, real-world images remain a crucial component for transferring the model to natural lighting distributions. Therefore, the final configuration employs a mixed training set that combines the controllability of synthetic scenes with the realism of actual lighting conditions. 

\subsection{Ablation Study Conclusions}

Ablation experiments demonstrate that the final quality is determined not by any single architectural complication, but by a combination of several relatively simple solutions. Relying on a zero LoRA state proves unreliable, as the base model does not establish a correct neutral point for the transformation in question. Combining lighting activation and deactivation into a single signed LoRA direction degrades the reverse operation, confirming the need to separate these tasks. Additional components in the reverse scheme are essential for establishing a stable editing direction, whereas excessively frequent normalization yields no further benefit. Finally, the inclusion of real-world images in the training set proves crucial for aligning light and shadow distributions when transferring the method to real-world scenes.

\section{Future Work}

Although the proposed approach demonstrates the feasibility of controllable artificial light editing using limited training data and compact LoRA adapters, the current solution leaves room for further development in several areas.

\subsection{Expanding the Training Dataset}

One of the most obvious avenues for improvement is increasing the size of the training dataset. In the current setup, the primary focus was on operating in a low-data regime; consequently, the adapters were trained on a limited set of carefully curated pairs.

Expanding the dataset could potentially enhance the model robustness across a wider variety of interiors, light source types, materials, and ambient lighting conditions. Increasing the number of real-world photographs is particularly important, as they better capture the complex interactions of light with surfaces, reflections, and shadows—elements that are not always fully replicated in synthetic scenes.

The balance between synthetic and real-world examples is also of particular interest. Although the researchers behind LightLab\cite{lightlab2025} addressed this issue extensively, the vast disparity in dataset sizes makes a direct application of their findings somewhat problematic. Synthetic data offers precise control over light source configurations and yields well-aligned pairs, whereas real-world images improve generalization to natural scene distributions. Future work will investigate the optimal ratio of these two data types.

\subsection{Transition to More Advanced Base Models}

Another avenue involves transitioning to more advanced generative image-editing models. The proposed method is not strictly tied to a specific base model architecture and can be adapted to newer families of image-editing models while retaining the general adapter training scheme.

Utilizing more advanced models could potentially improve the preservation of original scene details, enhance adherence to text instructions, and ensure more consistent modeling of complex lighting effects. Furthermore, advances in base models might reduce the need for separate auxiliary mechanisms if the model itself possesses a stronger prior regarding physically plausible lighting.

\subsection{Differentiating Base Models for Different Editing Directions}

The use of distinct base models for Light-On and Light-Off tasks is of particular interest. Experiments demonstrate that these operations differ significantly in nature; consequently, employing the same base architecture for both tasks is not necessarily optimal.

Moving forward, we plan to select a base model individually for each editing direction based on objective and perceptual metrics. Specifically, a promising approach involves using models from the Qwen family for the Light-On operation, while considering FLUX.2 as the base model for turning off the light source.

This separation aligns with the overall logic of the proposed method: just as Light-On and Light-Off require distinct training setups and separate adapters, they may similarly benefit from different base generative priors. In this context, the choice of architecture serves as an additional layer of system specialization.

Another area of interest is a scheme where adapters for models from different families are trained jointly and utilized as normalization components during the training of the opposite direction of the image-state change. This does not conflict with the separate task formulation, as each adapter is still optimized for its specific transformation direction; however, information derived from the model behavior during the reverse operation can serve as an additional regularization signal. Such a scheme could potentially facilitate better alignment between the two editing directions without reducing them to a single symmetric transformation or requiring the use of an identical base generative prior.

\subsection{Parametric Control of Light Source State}

In the current formulation, controlling the state of a light source is primarily framed as a discrete transition between On and Off states. However, continuous control of lighting parameters is of greater interest for practical applications.

As a next step, the plan is to incorporate the light source's brightness level of the light source into the control conditions. This will make it possible to specify not only whether the lamp is on or off but also the desired intensity of its illumination. Such a parameter can be applied to both Light-On and Light-Off transitions, transforming binary editing into a parametrically controlled change of the light source state.

Additionally, the model conditions are expected to include information about the type of shade, in particular, distinguishing between transparent and opaque designs. This parameter directly affects the nature of light propagation: a transparent shade usually produces a more locally pronounced glow and makes the source itself visually noticeable, whereas an opaque shade can redistribute the light more strongly and change the direction of its propagation.

Explicitly encoding these characteristics should reduce the ambiguity of text prompts and enable the model to generate more accurate distributions of brightness, shadows, and reflected light for various types of light fixtures.

\subsection{Outlook}

Thus, the further development of the method hinges primarily on expanding the volume and diversity of data, employing more powerful base models, and shifting from binary source state control to a more parametric problem formulation. Further specialization of base models for Light-On and Light-Off operations also appears promising, as it allows for accounting for the differences between generating a new light contribution and reconstructing the scene after its removal.

Collectively, these directions could lead to a more versatile artificial light editing system capable not only of toggling the state of a selected light source but also of controlling its intensity and accounting for the physical characteristics of the luminaire itself.

\section{Conclusion}

In this work, we have addressed the task of the controllable artificial light editing in indoor images and demonstrated that an effective solution does not necessarily require constructing a universal relighting model or using large-scale training datasets. Instead, we have proposed decomposing the original problem into two related yet fundamentally distinct tasks: turning a selected light source on and turning it off.

This distinction is one of the key concepts of the work. Despite the apparent symmetry of the operations, their underlying mechanisms differ: turning on a light involves adding a coherent lighting effect to an existing scene, whereas turning it off requires restoring the state that was previously altered by an active source. Experiments have demonstrated that attempting to combine both tasks into a single framework degrades performance in one of them. This confirms the necessity of treating these tasks separately—at the levels of training, adapter selection, and potentially the choice of base models.

Ablation studies have also highlighted the importance of specific components within the proposed scheme. Furthermore, the inclusion of real-world images in the training set have proved crucial for accurately reproducing light and shadow distributions.

The results obtained indicate that, given limited data, performance is determined less by the scale of the model or dataset than by the correct formulation of the task, the structure of the training signal, and the specialization of adapters for the specific type of scene state change. This approach enables controllable artificial light editing without significantly complicating the base architecture or requiring the training of a separate, universal relighting model. In practical terms, this reduces the burden of training data preparation and makes the proposed scheme more attractive in resource-constrained environments, as it does not require the creation of a large-scale, specialized dataset for adaptation.

The primary limitation of the current approach is its sensitivity to the training data distribution. The adapters struggle to generalize to light fixture types and source geometries that differ significantly from the examples in the training set. This points to a need for greater data diversity and a more systematic study of the adapter ability to generalize to previously unseen light source configurations. Future work will focus on investigating more advanced base image-editing models and exploring the use of different architectures based on their performance in specific tasks. A distinct area of research involves shifting from a binary source-state change to parametric control, incorporating parameters such as brightness levels and additional luminaire characteristics, for instance, the type and transparency of the light fixture's shade.

Overall, the proposed approach demonstrates that the task decomposition and the training of specialized LoRA adapters can serve as an effective foundation for the controllable editing of artificial lighting, particularly in data-constrained scenarios. The results also suggest that the future development of such systems may lie less in creating a more universal transformation model and more in the precise specialization of individual components for different types of scene-state changes.

\clearpage

\bibliographystyle{IEEEtran}

\clearpage

\appendix
\section{Comparison of Results for Light On}
\label{app:light-activation}

This Appendix presents examples of our method operation as compared with other methods for the task of turning on the light. Each row contains the original image, the result obtained by our method, and the result obtained by a competing method, in that order.

\begin{figure}[!htbp]
  \centering
  \includegraphics[
    width=\linewidth,
    height=0.83\textheight,
    keepaspectratio
  ]{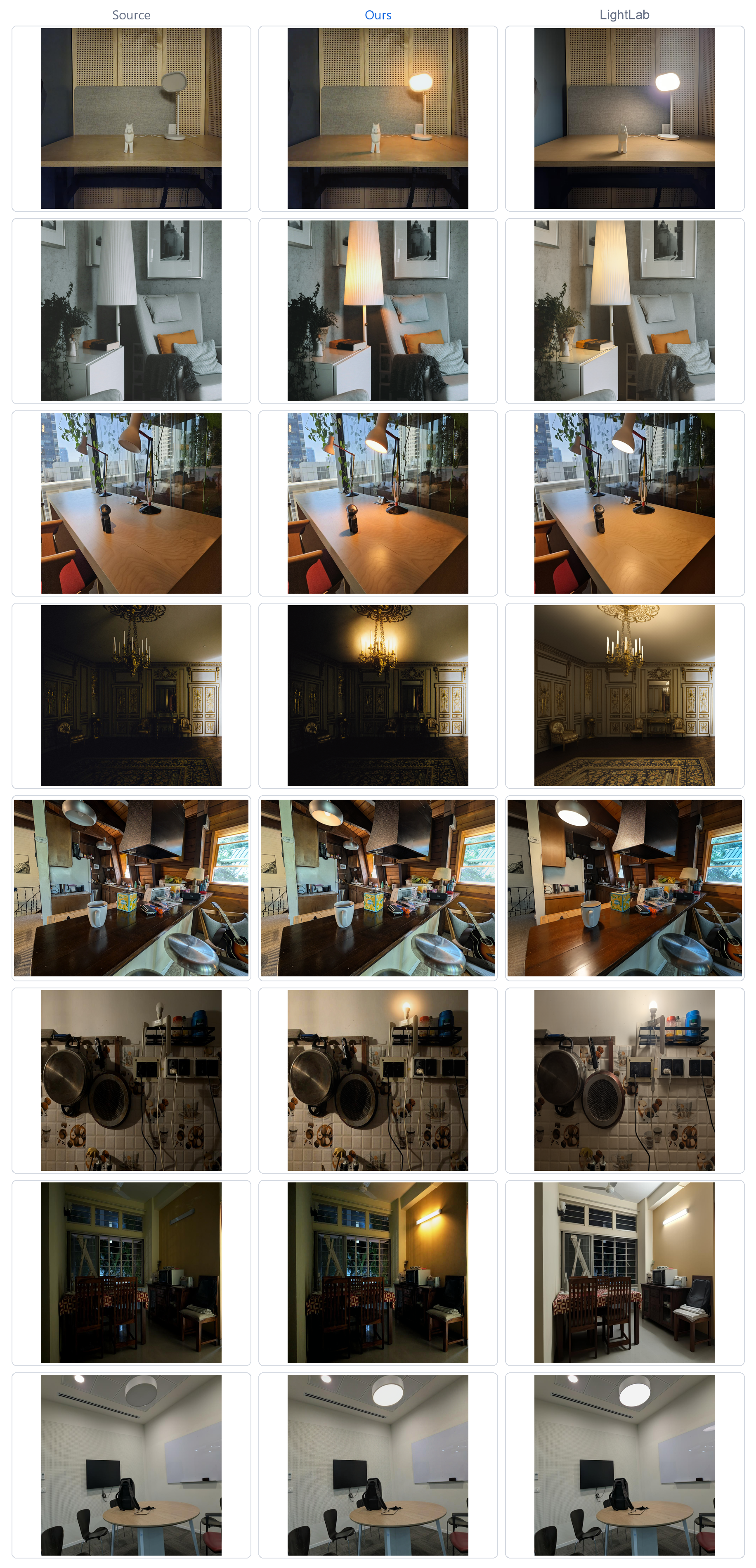}
  \caption{Qualitative comparison between our method and LightLab
  for light activation.}
  \label{fig:appendix-light-on-lightlab}
\end{figure}
\clearpage

\begin{figure*}[p]
  \centering
  \includegraphics[width=\textwidth,height=0.84\textheight,keepaspectratio]{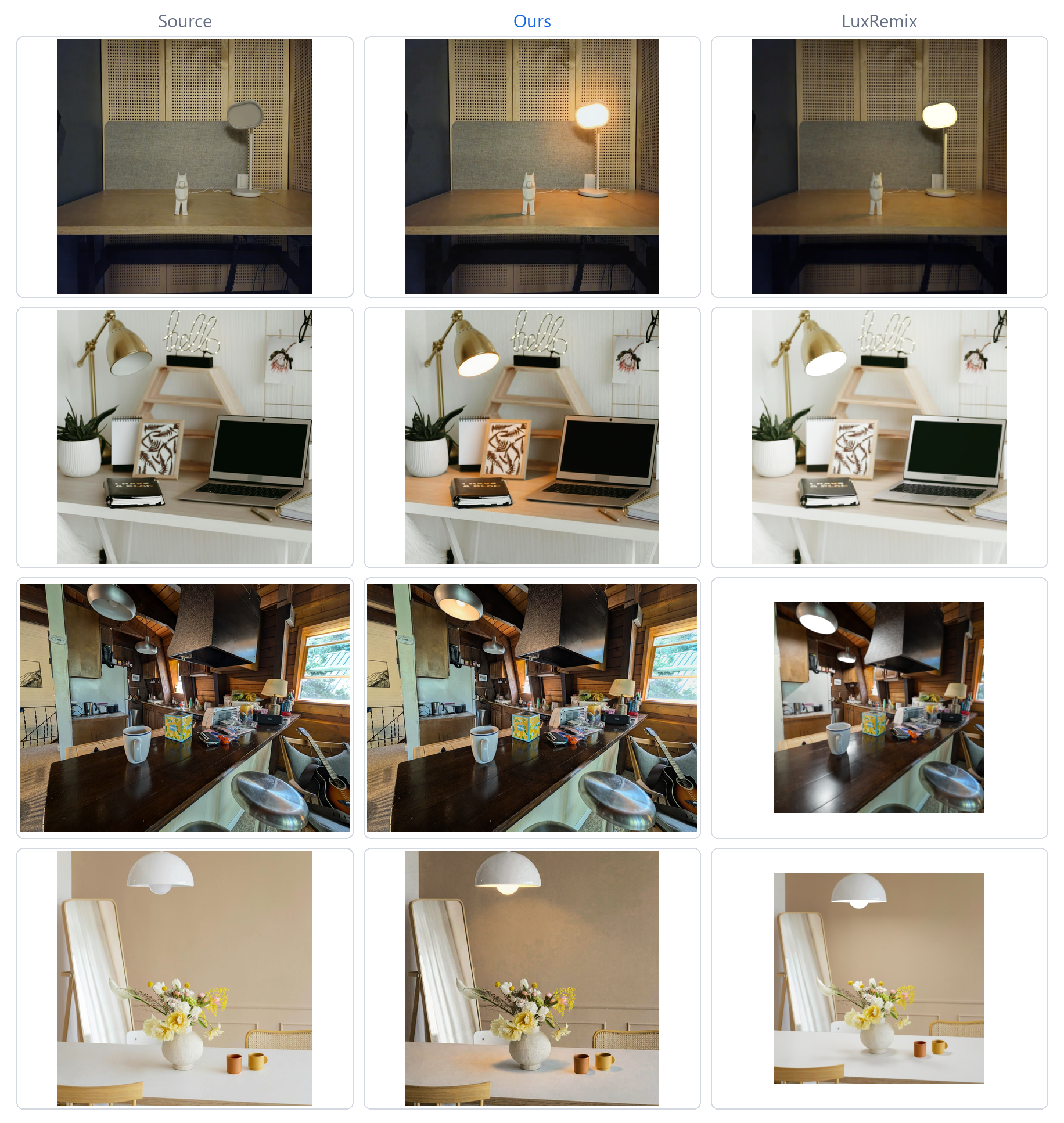}
  \caption{Qualitative comparison between our method and LuxRemix for light activation.}
  \label{fig:appendix-light-on-luxremix}
\end{figure*}
\clearpage

\begin{figure*}[p]
  \centering
  \includegraphics[width=\textwidth,height=0.84\textheight,keepaspectratio]{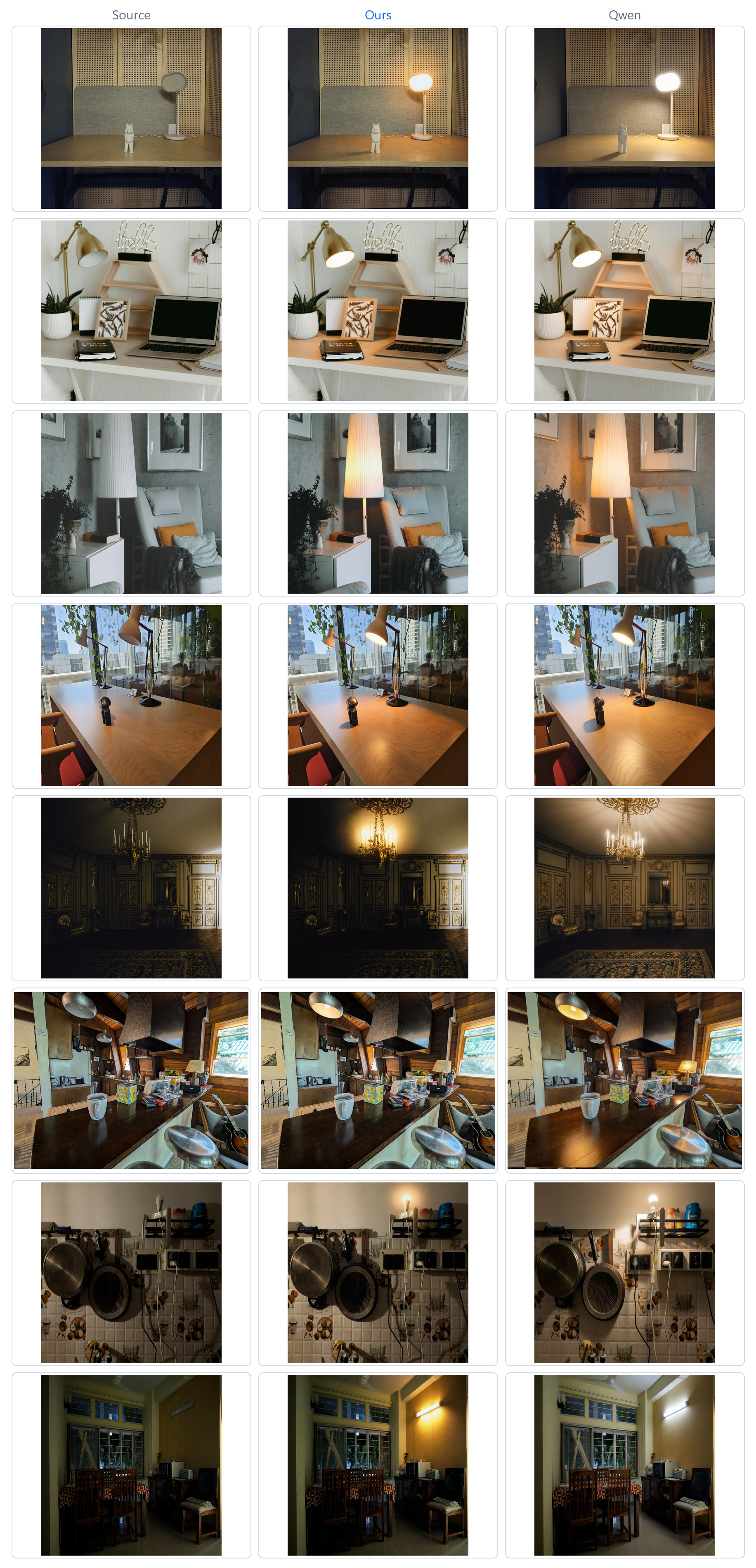}
  \caption{Qualitative comparison between our method and Qwen for light activation.}
  \label{fig:appendix-light-on-qwen}
\end{figure*}
\clearpage

\begin{figure*}[p]
  \centering
  \includegraphics[width=\textwidth,height=0.84\textheight,keepaspectratio]{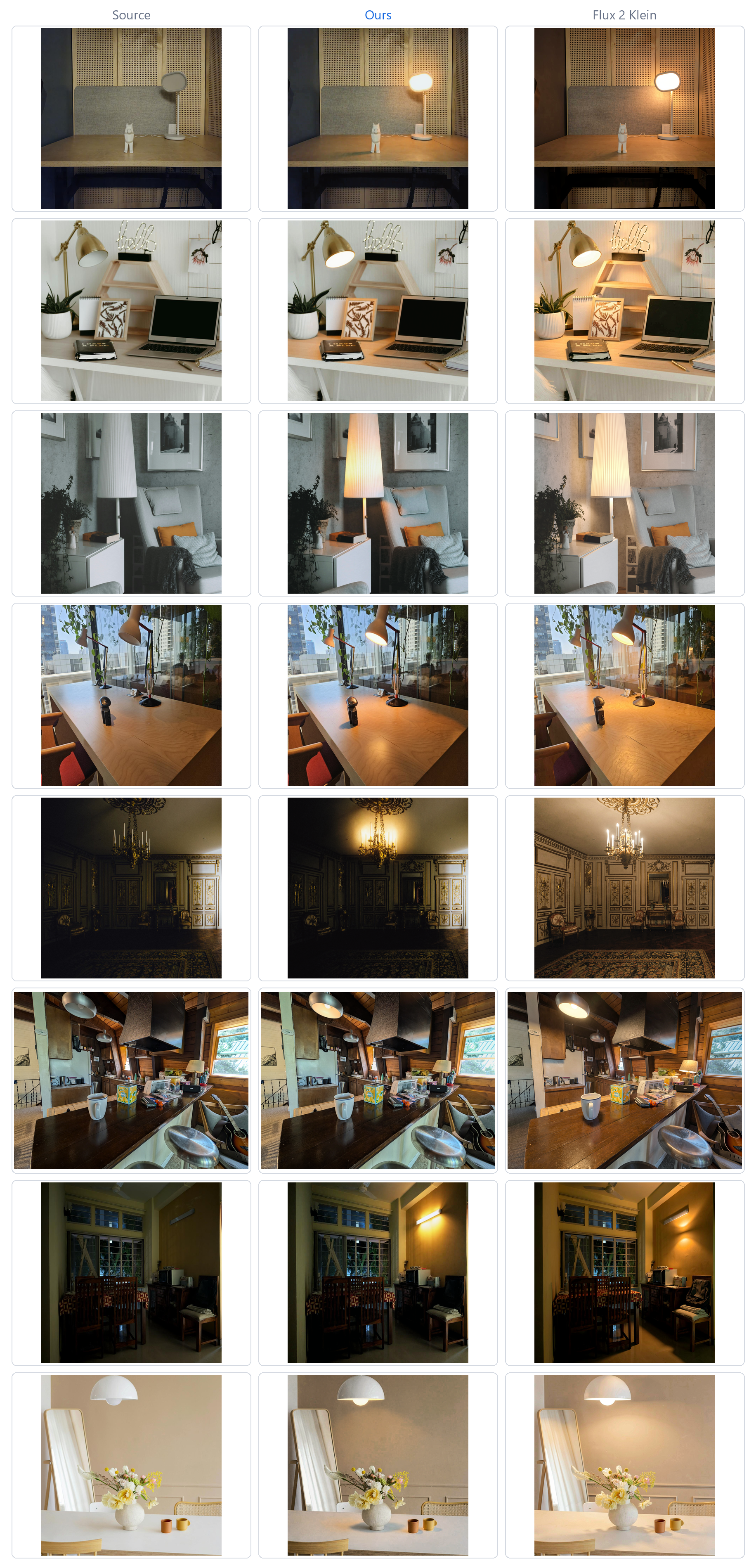}
  \caption{Qualitative comparison between our method and Flux~2~Klein for light activation.}
  \label{fig:appendix-light-on-flux-klein}
\end{figure*}
\clearpage


\section{Comparison of Results for Light Off}
\label{app:light-activation}

This Appendix presents examples demonstrating the performance of our method as compared to other methods for the Light-Off task. Each row displays the original image, the result obtained by our method, and the result obtained by a competing method, in that order.

\begin{figure}[!htbp]
  \centering
  \includegraphics[width=\linewidth,height=0.72\textheight,keepaspectratio]{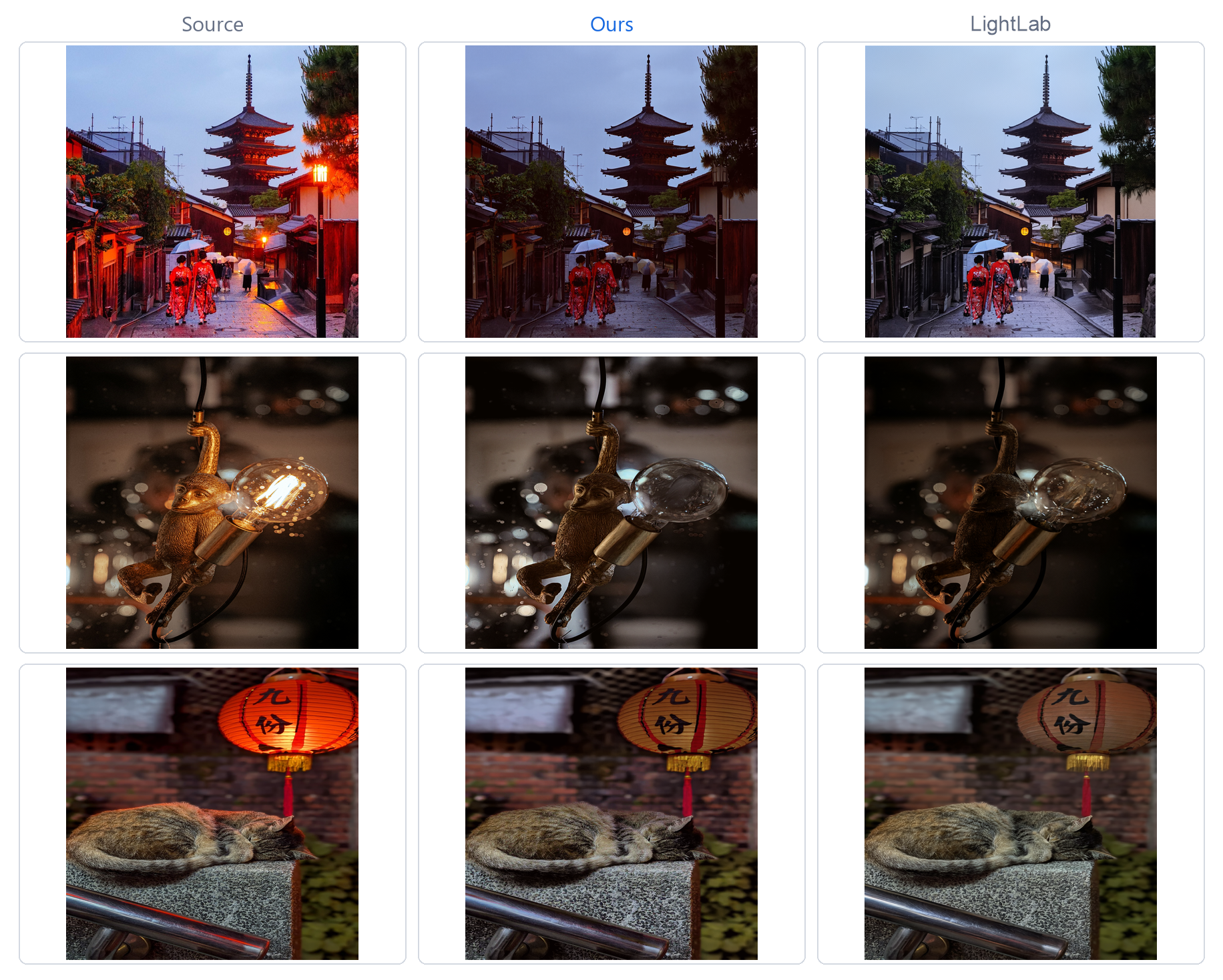}
  \caption{Qualitative comparison between our method and LightLab for light deactivation.}
  \label{fig:appendix-light-off-lightlab}
\end{figure}

\begin{figure}[!htbp]
  
  \includegraphics[width=\linewidth,height=0.72\textheight,keepaspectratio]{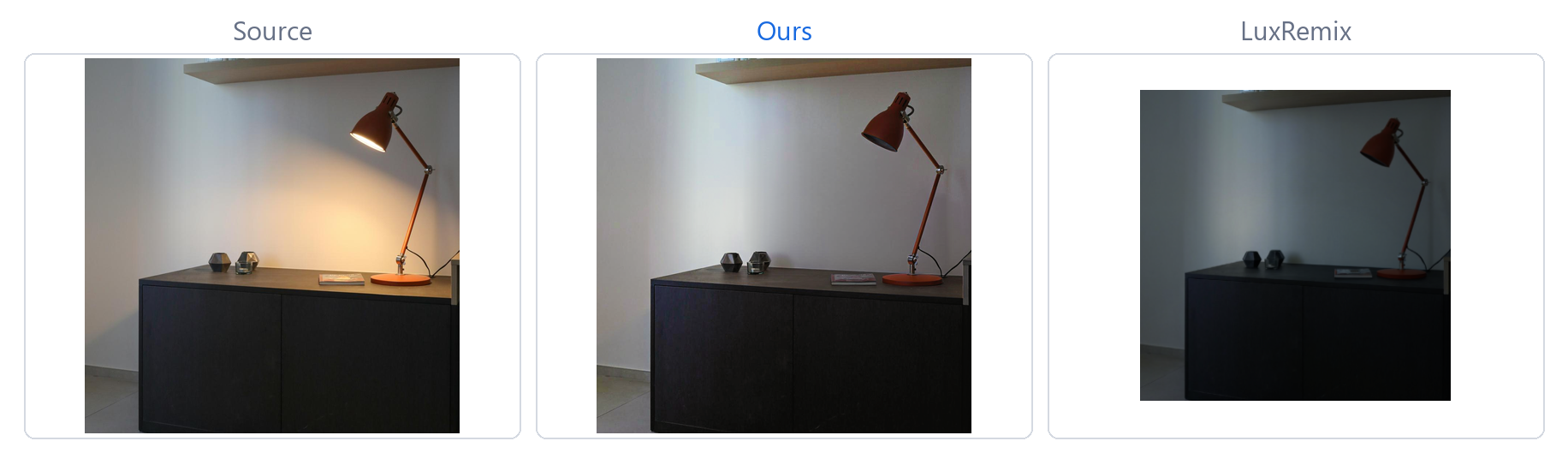}
  \caption{Qualitative comparison between our method and LuxRemix for light deactivation.}
  \label{fig:appendix-light-off-luxremix}
\end{figure}

\begin{figure}[!htbp]
  \centering
  \includegraphics[width=\linewidth,height=0.78\textheight,keepaspectratio]{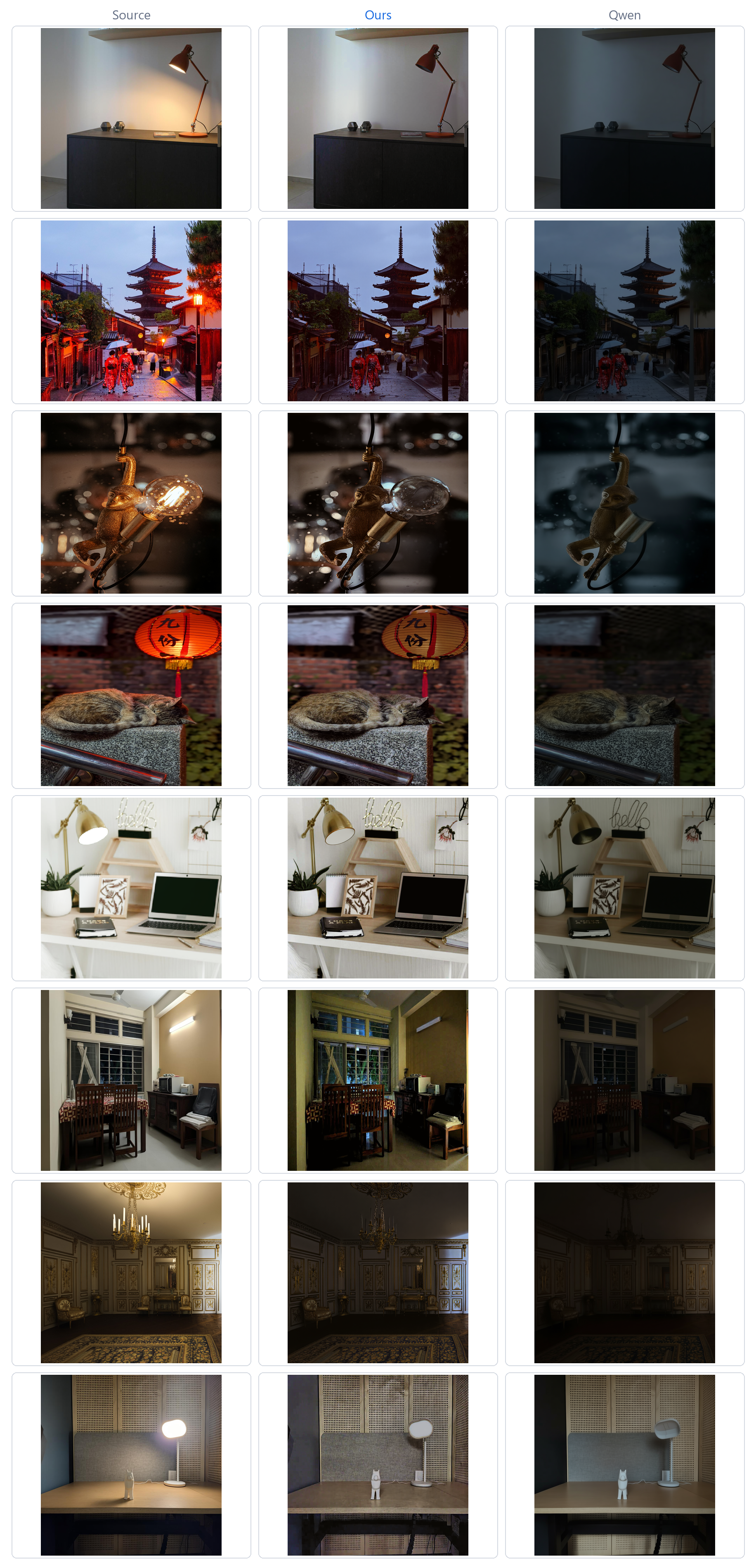}
  \caption{Qualitative comparison between our method and Qwen for light deactivation.}
  \label{fig:appendix-light-off-qwen}
\end{figure}

\begin{figure}[!htbp]
  \centering
  \includegraphics[width=\linewidth,height=0.78\textheight,keepaspectratio]{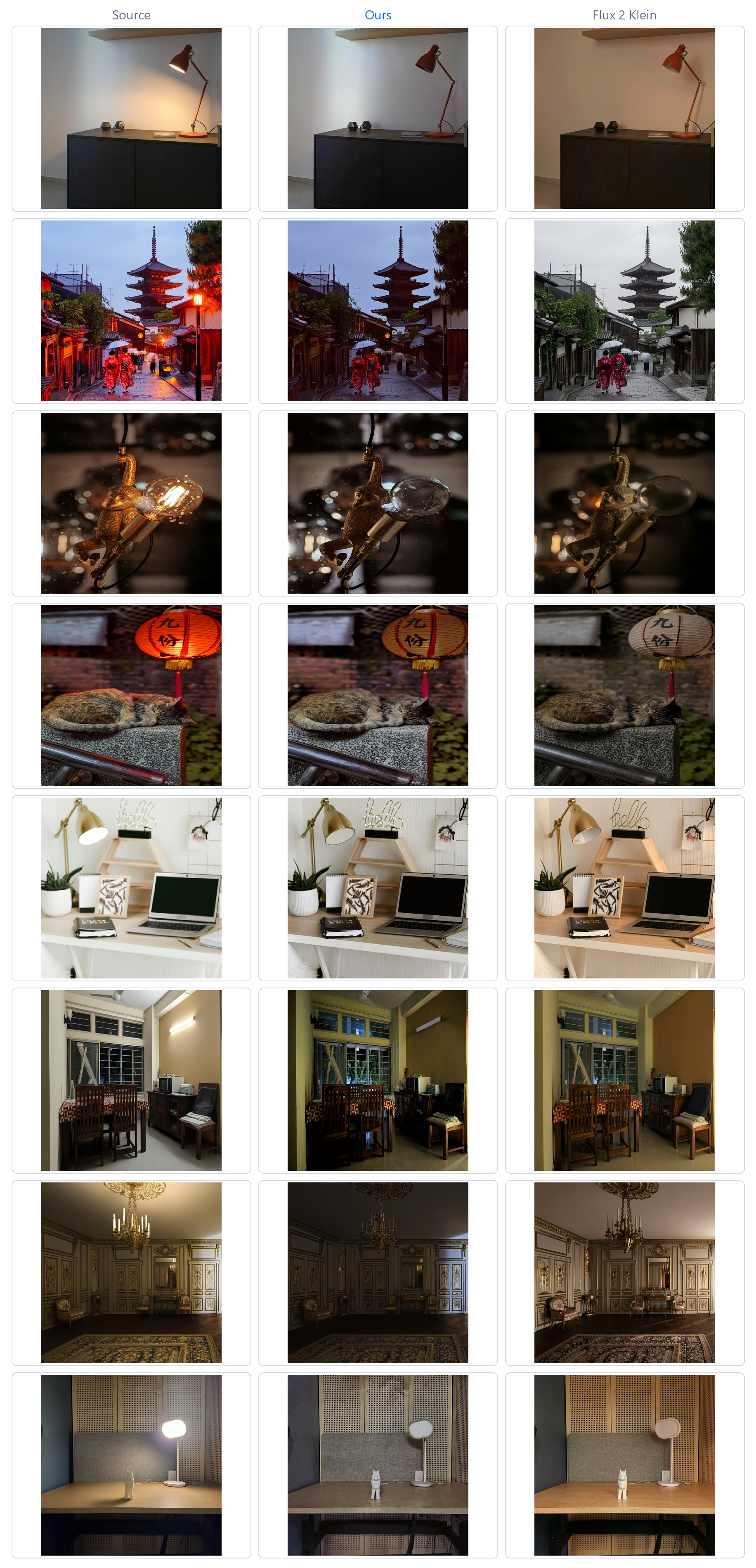}
  \caption{Qualitative comparison between our method and Flux~2~Klein for light deactivation.}
  \label{fig:appendix-light-off-flux-klein}
\end{figure}

\clearpage

\end{document}